\pdfoutput=1
\documentclass[11pt]{article}

\usepackage[final]{acl}
\usepackage{times}
\usepackage{latexsym}

\newenvironment{myquote}{\list{}{\leftmargin=0.2in\rightmargin=0.2in}\item[]}%
  {\endlist}

\usepackage[T1]{fontenc}
\usepackage[utf8]{inputenc}

\usepackage{microtype}

\usepackage{inconsolata}

\usepackage{amsmath,amssymb}

\usepackage{multirow}
\usepackage{tabularx}
\usepackage{booktabs}
\usepackage{xcolor,caption,subcaption,graphicx}
\usepackage{listings}

\title{Ranking Entities along Conceptual Space Dimensions with LLMs:\\ An Analysis of Fine-Tuning Strategies}

\author{Nitesh Kumar \and Usashi Chatterjee \and Steven Schockaert\\
  Cardiff NLP, School of Computer Science and Informatics \\
  Cardiff University, United Kingdom \\
  \texttt{\{kumarn8,chatterjeeu,schockaerts1\}@cardiff.ac.uk} \\}

\begin{document}
\maketitle
\begin{abstract}
Conceptual spaces represent entities in terms of their primitive semantic features. Such representations are highly valuable but they are notoriously difficult to learn, especially when it comes to modelling perceptual and subjective features. Distilling conceptual spaces from Large Language Models (LLMs) has recently emerged as a promising strategy, but existing work has been limited to probing pre-trained LLMs using relatively simple zero-shot strategies. We focus in particular on the task of ranking entities according to a given conceptual space dimension. Unfortunately, we cannot directly fine-tune LLMs on this task, because ground truth rankings for conceptual space dimensions are rare. We therefore use more readily available features as training data and analyse whether the ranking capabilities of the resulting models transfer to perceptual and subjective features. We find that this is indeed the case, to some extent, but having at least some perceptual and subjective features in the training data seems essential for achieving the best results.\footnote{Our pre-processed datasets and code can be found at \url{https://github.com/niteshroyal/RankingUsingLLMs}.} 
%We also find that pointwise ranking strategies are competitive against pairwise approaches, in defiance of common wisdom.
\end{abstract}

\section{Introduction}
Knowledge graphs (KGs) have emerged as the \emph{de facto} standard for representing knowledge in areas such as Natural Language Processing \cite{schneider-etal-2022-decade}, Recommendation \cite{DBLP:journals/tkde/GuoZQZXXH22} and Search \cite{DBLP:journals/ftir/ReinandaMR20}.  However, much of the knowledge that is needed in applications is about \emph{graded} properties, e.g.\ recipes being healthy, movies being original or cities being kids-friendly. Such knowledge is easiest to model in terms of rankings: we can rank recipes according to how healthy they are even if we cannot make a hard decision about which ones are healthy and which ones are not. For this reason, we argue that conceptual spaces \cite{DBLP:books/daglib/0006106} should be used, alongside knowledge graphs, in many settings. 

A conceptual space specifies a set of quality dimensions, which correspond to primitive semantic features. For instance, in a conceptual space of movies, we might have a quality dimensions reflecting how original a movie is. Entities are represented as vectors, specifying a suitable feature value for each quality dimension. While the framework of conceptual spaces is more general, we will essentially view quality dimensions as rankings. 

Conceptual spaces have the potential to play a central role in various knowledge-intensive applications. In the context of recommendation, for instance, they could clearly complement the factual knowledge that is captured by typical KGs (e.g.\ modelling the style of a movie, rather than who directed it), making it easier to infer user preferences from previous ratings. They could also be used to make recommendations more controllable, as in the case of critiquing-based systems, allowing users to specify feedback of the form ``like this movie, but more kids-friendly'' \cite{DBLP:journals/umuai/ChenP12,DBLP:journals/tiis/VigSR12}. Conceptual spaces furthermore serve as a natural interface between neural and symbolic representations \cite{DBLP:journals/ai/AisbettG01}, and may thus enable principled explainable AI methods.

However, the task of learning conceptual spaces has proven remarkably challenging. The issue of reporting bias \cite{DBLP:conf/cikm/GordonD13}, in particular, has been regarded as a fundamental obstacle: the knowledge captured by conceptual spaces is often so obvious to humans that it is rarely stated in text. For instance, the phrase ``green banana'' is more frequent in text than ``yellow banana'' \cite{paik-etal-2021-world}, as the colour is typically not specified when yellow bananas are discussed. \citet{paik-etal-2021-world} found that predictions of Language Models (LMs) about the colour of objects were correlated with the distribution of colour terms in text corpora, rather than with human judgements, suggesting that LMs cannot overcome the challenges posed by reporting bias. However, \citet{liu-etal-2022-ever} found that larger LMs can perform much better on this task. Going beyond colour, \citet{chatterjee-etal-2023-cabbage} evaluated the ability of LLMs to predict taste-related features, such as sweetness and saltiness, obtaining mixed results: the rankings predicted by LLMs, in a zero-shot setting, had a reasonable correlation with human judgments but they were not consistently better than those produced by a fine-tuned BERT \cite{devlin-etal-2019-bert} model. 

%This clearly suggests that standard zero (or few) shot probing techniques are not sufficient for extracting conceptual space representations.

In this paper, we analyse whether LLMs can be fine-tuned to extract better conceptual space representations. The difficulty is that ground truth rankings are typically not available when it comes to perceptual and subjective features, outside a few notable exceptions such as the aforementioned taste dataset. We therefore explore whether more readily available features can be used for fine-tuning the model. For instance, we can obtain ground truth rankings from Wikidata entities with numerical attributes (e.g.\ the length of rivers, the birth date of people, or the population of cities) and then use these rankings to fine-tune an LLM. We furthermore compare two different strategies for ranking entities with an LLM: the \emph{pointwise} approach uses an LLM to assign a score to each entity, given some feature, while the \emph{pairwise} approach uses an LLM to decide which among two given entities has the feature to the greatest extent. Our contributions and findings can be summarised as follows:
\begin{itemize}
\item We evaluate on three datasets which have not previously been used for studying language models: a dataset of rocks, a dataset of movies and books, and a dataset about Wikidata entities. We use these datasets alongside datasets about taste \cite{chatterjee-etal-2023-cabbage} and physical properties \cite{li-etal-2023-language-models}.
\item We analyse whether fine-tuning LLMs on features from one domain (e.g.\ taste) can improve their ability to rank entities in different domains (e.g.\ rocks). We find this indeed largely to be the case, as long as the training data also contains perceptual or subjective features.
\item We compare pointwise and pairwise approaches for ranking entities with LLMs. Despite the fact that pairwise approaches have consistently been found superior for LLM-based document ranking \cite{DBLP:journals/corr/abs-1910-14424,DBLP:conf/ictir/GienappFHP22,DBLP:journals/corr/abs-2306-17563}, when it comes to comparing entities, we find the pointwise approach to be highly effective, although the best results overall are still obtained using a pairwise model.
\item To obtain \emph{rankings} from pairwise judgments, we need a suitable strategy for aggregating these judgments. %We analyse the effectiveness of two principled aggregation methods. The best results 
We show the effectiveness of an SVM based strategy for this purpose. While this strategy is known to have desirable theoretical properties, it has not previously been considered in the context of language models, to the best of our knowledge
\end{itemize}

%and evaluate whether the resulting models generalise to unseen attributes, and to perceptual and subjective features in particular. We experiment with both pointwise and pairwise ranking strategies. 

%****************************************************
\section{Related Work}

\paragraph{LMs as Knowledge Bases} Our focus in this paper is on extracting knowledge from language models. This idea of language models as knowledge bases was popularised by \citet{petroni-etal-2019-language}, who showed that the pre-trained BERT model captures various forms of factual knowledge, which can moreover be extracted using a simple prompt. Work in this area has focused on two rather distinct goals. On the one hand, probing tasks, such as the one proposed by \citet{petroni-etal-2019-language}, have been used as a mechanism for analysing and comparing different language models. On the other hand, extracting knowledge from LMs has also been studied as a practical tool for building or extending symbolic knowledge bases. This has been particularly popular for capturing types of knowledge which are not commonly found in traditional knowledge bases, such as commonsense knowledge \cite{bosselut-etal-2019-comet,west-etal-2022-symbolic,yu-etal-2023-folkscope}. Several works have focused on distilling KGs from language models \cite{cohen-etal-2023-crawling}. \citet{hao-etal-2023-bertnet} studied this problem for non-traditional relations such as ``is capable of but not good at''. Along the same lines, \citet{DBLP:journals/corr/abs-2305-15002} have focused on modelling relations that are a matter of degree, such as ``is a competitor of'' or ``is similar to''. We can similarly think of the conceptual space dimensions that we consider in this paper as gradual properties. 

Where the aforementioned approaches explicitly extract knowledge from an LM, the knowledge captured by LMs has also been used implicitly, by applying such models in a wide range of knowledge-intensive applications, including 
%\usashi{avoid bias reporting \cite{shwartz-choi-2020-neural}}, 
closed-book question answering \cite{roberts-etal-2020-much}, knowledge graph completion \cite{DBLP:journals/corr/abs-1909-03193}, recommendation \cite{DBLP:conf/cikm/SunLWPLOJ19,DBLP:conf/recsys/Geng0FGZ22}, entity typing \cite{huang-etal-2022-unified} and ontology alignment \cite{DBLP:conf/aaai/0008CA022}, to name just a few. 

\paragraph{Conceptual Space of LMs} 
There is an ongoing debate about the extent to which LMs can truly capture meaning \cite{bender-koller-2020-climbing,abdou-etal-2021-language,patel2022mapping,DBLP:journals/mima/Sogaard23}. Within this context, several authors have analysed the ability of LMs to predict perceptual features. As already mentioned, \citet{paik-etal-2021-world} and \citet{liu-etal-2022-ever} analysed the ability of LMs to predict colour terms. \citet{abdou-etal-2021-language} analysed whether the representation of colour terms in LMs can be aligned with their representation in the CIELAB colour space. \citet{patel2022mapping} similarly showed that LLMs can generate colour terms from RGB codes in a few-shot setting, even if the codes represent a rotation of the standard RGB space. They also show a similar result for terms describing spatial relations. \citet{zhu2024recovering} have similarly shown that LLMs can understand colour codes, by using them to generate HSL codes for everyday objects, or by asking models to choose the most suitable code among two alternatives. 

Beyond the colour domain, \citet{li-etal-2023-language-models} considered physical properties such as height or mass. While they found LLMs to struggle with such properties, \citet{chatterjee-etal-2023-cabbage} reported better results on the same datasets, especially when using GPT-4. Focusing on visual features, \citet{DBLP:conf/iclr/MerulloCEP23} showed that the representations of concepts in vision-only and text-only models can be aligned using a linear mapping.
\citet{chatterjee-etal-2023-cabbage} focused on the taste domain, modelling properties such as sweet. They found that GPT-3 can model such properties to a reasonable extent, but not better than a fine-tuned BERT model.

\citet{gupta-etal-2015-distributional} already considered the problem of modelling gradual properties in the context of static word embeddings, although their analysis was limited to objective numerical features. \citet{DBLP:journals/ai/DerracS15} similarly learned conceptual space dimensions for properties such as ``violent'' in a semantic space of movies, while \citet{rubinstein-etal-2015-well} assessed the ability to predict (gradual) commonsense properties from word embeddings. More recently, \citet{grand2022semantic} also used linear projections to map word embeddings to dimensions capturing gradual feature (e.g.\ size). \citet{erk2024adjusting} argue for the importance of using human ratings to learn such projections.
%employed seed based vectors along with human rating to predict object properties (size, danger, etc) and stylistic properties (formality and complexity). 
Note that all these approaches essentially learn a linear classifier or regression model for each property independently, and can thus not generalise to new properties. A related line of work has focused on modelling the rankings associated with scalar adverbs \cite{lorge2023wacky} and adjectives \cite{samir-etal-2021-formidable}. 
%****************************************************
\section{Extracting Rankings}
We consider the following problem: given a set of entities $\mathcal{E}$ and a feature $f$, rank the entities in $\mathcal{E}$ according to the their value for the feature $f$. In some cases, $f$ will refer to a numerical attribute. For instance, $\mathcal{E}$ may be a set of countries and $f$ the population of a country, where the task is then to rank the countries according to their population. In other cases, $f$ will rather refer to a gradual property. For instance, $\mathcal{E}$ may be a set of food items and $f$ may be the level of sweetness. Let us write $f(e)$ for the value of feature $f$ for entity $e$. 

We consider two broad strategies for solving the considered ranking task with LLMs. First, we can use LLMs to map each entity $e$ to some score $w(e)$, with the assumption that $w(e_1) < w(e_2)$ iff $f(e_1) < f(e_2)$. This \emph{pointwise} approach to learning to rank is considered in Section \ref{secPointwise}. Second, we can use LLMs to solve a binary classification problem: given two entities $e_1$ and $e_2$, decide whether $f(e_1)<f(e_2)$ holds. This \emph{pairwise} approach needs to be combined with a strategy for aggregating the LLM predictions into a single ranking. The main disadvantage is that a large number of judgements need to be collected for this to be effective, which means that such approaches are less efficient than pointwise strategies. However, in the context of document retrieval, pairwise approaches have been found to outperform pointwise approaches \cite{DBLP:journals/corr/abs-1910-14424,DBLP:conf/ictir/GienappFHP22,DBLP:journals/corr/abs-2306-17563}.
We discuss pairwise and pointwise strategies in Sections \ref{secPairwise} and \ref{secPointwise} respectively. Finally, Section \ref{secBaselines} describes how we establish baseline results using ChatGPT and GPT4.

%*******************
\subsection{Pairwise Model}\label{secPairwise}
The problem of predicting whether $f(e_1) < f(e_2)$ holds can be straightforwardly cast as a sequence classification problem. To this end, we use a prompt of the following form:
\begin{myquote}
This question is about two \textsc{[entity type]}: [Is/Does/Was] \textsc{[entity 1]} \textsc{[comparative feature]} than \textsc{[entity 2]}?
\end{myquote}
Note that the exact formulation depends on the type of feature which is used for ranking.
For instance, some instantiations of the prompt are as follows: 
\begin{itemize}
\item \textit{This question is about two rivers: Is River Thames longer than Seine?}
\item \textit{This question is about two companies: Was Meta founded after Alphabet?}
\item \textit{This question is about two food items: Does banana taste sweeter than chicken?}
\end{itemize}
In initial experiments, we used prompts with a more uniform style (e.g.\ ``should \textsc{[entity 1]} be ranked higher than \textsc{[entity 2]} in terms of \textsc{[feature]}''). However, this inevitably leads to less natural sounding prompts for certain features, which may affect performance. Moreover, such prompts were sometimes found to be ambiguous (e.g.\ does ``ranked higher in terms of date of birth'' mean younger people should be ranked highest?).  
To obtain judgments about entity pairs, we use a standard sequence classification approach, where a linear layer with sigmoid activation is applied to the final hidden state. The model is trained using binary cross entropy using a set of training examples. % (see Section \ref{secDatasets}).

\paragraph{Aggregating judgments}
We typically want to rank a given set of entities, rather than judging the relative position of two particular elements. This means that we need a strategy for aggregating (noisy) pairwise judgments into a single ranking. This problem has received extensive attention in the literature, where standard techniques include spectral ranking \cite{vigna2016spectral} and maximum likelihood estimation w.r.t.\ an underlying statistical model. However, existing approaches often consider a stochastic setting, where we may have access to several judgments for the same entity pair (e.g.\ when ranking sports teams based on the outcomes of head-to-head matches). 

Our setting is slightly different, as we can realistically only obtain judgments for a small sample of entity pairs. In particular, we ideally need methods with $\Omega(n)$ sample complexity, i.e.\ methods that can perform well with a number of judgements that is linear in the number of queries. \citet{DBLP:conf/icml/WauthierJJ13} discuss two such methods. Let us write $e_1,...,e_n$ for the entities to be ranked. The first method uses a linear SVM to learn a weight vector $\mathbf{w}=(w_1,...,w_n)$.  Let $\mathbf{x}_i$ be an $n$-dimensional one-hot vector, which is $1$ in the $i$\textsuperscript{th} coordinate and 0 elsewhere. If we have a pairwise judgement $f(e_i) < f(e_j)$ then this is translated into the constraint that $\mathbf{w}(\mathbf{x}_j-\mathbf{x}_i)>0$. A standard \textbf{SVM} can then be used to find the vector $\mathbf{w}$ that maximises the margin between positive and negative examples. Entity $e_i$ is ranked based on its corresponding weight $w_i$. The second method simply scores each entity $e_i$ based on the number of pairwise comparisons where $e_i$ was ranked higher/lower. Specifically, let us define $s_{ij}=1$ if entities $e_i$ and $e_j$ have been compared, and $s_{ij}=0$ otherwise. Furthermore, we define $c_{ij}=1$ if $f(e_i)>f(e_j)$, according to a pairwise comparison that was made, and $c_{ij}=-1$ otherwise. 
Then we can choose the weights as:
\begin{align*}
w_i = \frac{\sum_{j\neq i} s_{ij} c_{ij}}{\sum_{j\neq i} s_{ij}}
\end{align*}
We will refer to this strategy as \textbf{Count}. 
%Finally, we also consider a variant of this strategy which takes into account the confidence of the LLM classifier in each pairwise judgement, which we will refer to as \textbf{Weighted}. In this case, we still use \eqref{eqBordaCount}, but instead we define $c_{ij}$ to be the confidence of the classifier.

% pairwise document ranking: https://dl.acm.org/doi/abs/10.1145/3539813.3545140
%https://arxiv.org/pdf/2306.17563.pdf

% listwise document ranking:
% https://arxiv.org/pdf/2304.09542.pdf

% SVM setting for case with lots of observations: http://proceedings.mlr.press/v32/rajkumar14.pdf

% Rank centrality: https://arxiv.org/pdf/1209.1688.pdf

% Active ranking: https://teams.microsoft.com/l/message/19:5feceb2459154ac18585961849fdc301@thread.v2/1706114793354?context=%7B%22contextType%22%3A%22chat%22%7D

% Justification for Borda:
% https://www.jmlr.org/papers/volume18/16-206/16-206.pdf
% http://proceedings.mlr.press/v28/wauthier13.pdf (ALSO mentions SVM method)

%*******************
\subsection{Pointwise Model}\label{secPointwise}
For the pointwise model, we need to learn a scoring function $w: \mathcal{E}\rightarrow \mathbb{R}$. To this end, we use a prompt of the following form:
%In theoretical work on rank aggregation \cite{DBLP:journals/jmlr/ShahW17}, the ``quality'' of an entity is often modelled in terms of the probability that it wins a comparison from a randomly sampled entity. Inspired by this, we define the target $t_i$ of entity $e_i$ as its percentile in the ground truth ranking, e.g.\ $t_i=1$ if $e_i$ is the top-ranked entity, $t_i=0$ if $e_i$ is the bottom-ranked entity and $t_i=0.5$ if $e_i$ occurs precisely in the middle of the ranking. To train the scoring function, we use the following prompt:
\begin{myquote}
Is \textsc{[entity 1]} among \textsc{[superlative feature]} \textsc{[entity type]}?
\end{myquote}
For instance, \emph{Is River Thames among the longest rivers?} For each entity $e_i$, we obtain a score $w(e_i)\in\mathbb{R}$ by applying a linear layer to the final hidden state. Intuitively, $w(e_i)$ captures the (latent) quality of $e_i$ w.r.t.\ the considered feature. Since we cannot obtain ground truth labels for this score, we again rely on pairwise comparisons for training the model. Specifically, we estimate the probability $p_{ij}$ that  $f(e_i)>f(e_j)$ holds as:
$$
p_{ij} = \sigma\big(w(e_i)-w(e_j)\big)
$$
Then we use binary cross entropy as follows:
$$
\mathcal{L} = -\Big(\sum_{i\neq j} t_{ij} \log p_{ij} + (1-t_{ij})\log(1-p_{ij})\Big)
$$
where $t_{ij}=1$ if $f(e_i)>f(e_j)$ and $t_{ij}=0$ otherwise, and the summation ranges over all distinct entity pairs $e_i,e_j$ within the given mini-batch. Note that while we use pairwise comparisons for training the model, it is still a pointwise approach as the model produces scores for individual entities.

%*******************
\subsection{Baselines}\label{secBaselines}
To put the performance of the fine-tuning strategies from Sections \ref{secPairwise} and \ref{secPointwise} into context, we compare them with two conversational models: ChatGPT (\text{gpt-3.5-turbo}) and GPT-4 (\texttt{gpt-4}). We use both models in a zero-shot setting. For this purpose, we use the same prompt as in Section \ref{secPairwise} but append the sentence \emph{Only answer with yes or no.} Despite this instruction, the models occasionally still generated a different response, typically expressing that the question cannot be answered. For such entity pairs, we replace the generated response with a randomly generated label (\textit{yes} or \textit{no}).\footnote{Statistics about how often this was needed can be found in the appendix.} 

%Second, we also experiment with \textit{davinci-002} and \textit{babbage-002}. For these models, the OpenAI API allows us to obtain conditional probabilities, which is no longer possible with newer models. Specifically, we can score entities using the following prompt
%\begin{myquote}
%The following question is about \textsc{[entity type]}. In terms of \textsc{[feature]}, \textsc{[entity]} should be ranked
%\end{myquote}
%Then we use the probability that the next token is \emph{high} to determine the rank of the entity. 

%\begin{myquote}
%The following question is about \textsc{[entity type]}. Should \textsc{[entity 1]} be ranked higher than \textsc{[entity 2]} in terms of \textsc{[feature]}?
%\end{myquote}

%****************************************************
\section{Datasets}\label{secDatasets}
In our experiments, we will rely on the following datasets, either for training or for testing the models. Each dataset consists of a number of rankings, where each ranking is defined by a set of entities and a feature along which the entities are ranked.

\paragraph{Wikidata} We have obtained 20 rankings from numerical features that are available on Wikidata\footnote{\url{https://www.wikidata.org/wiki/Wikidata:Main_Page}}. For instance, we obtained a ranking of rivers by comparing their length.\footnote{The entity types and corresponding features are listed in the appendix.} If there were more than 1000 entities with a given feature value, we selected the most 1000 popular entities. To estimate the popularity of an entity, we use their QRank\footnote{\url{https://qrank.wmcloud.org}}, which counts the number of page views of the corresponding entry in sources such as Wikipedia. For the entity type person, we limited the analysis to people born in London (which made it possible to retrieve the required information from Wikidata more efficiently). We similarly only considered museums located in Italy. For some experiments, we split the collected data in two datasets, called \textbf{WD1} and \textbf{WD2}. This will allow us to test whether models trained on one set of features (i.e.\ WD1) generalise to a different set of features (i.e.\ WD2). WD1 contains rankings which were cut off at 1000 elements, whereas WD2 contains rankings with fewer elements. 
%For both datasets, we randomly selected 80\% of the entities in each ranking to be used for training, and we test models on the remaining 20\% of the entities. 
We will write \textbf{WD} to refer to the full dataset, i.e.\ WD1 and WD2 combined. 
%We call the resulting datasets \textbf{Wikidata-1-train}, \textbf{Wikidata-1-test}, \textbf{Wikidata-2-train} and \textbf{Wikidata-2-test}.

\begin{table*}[t]
\fontsize{7.7pt}{9.24pt}\selectfont
\centering
\setlength\tabcolsep{3pt}
\begin{tabular}{l cc cccccc ccccccc cc ccc c}
\toprule
& \multicolumn{2}{c}{\textbf{Wikidata}} & \multicolumn{6}{c}{\textbf{Taste}} & \multicolumn{7}{c}{\textbf{Rocks}} & \multicolumn{2}{c}{\textbf{TG}} & \multicolumn{3}{c}{\textbf{Phys}}\\
\cmidrule(lr){2-3}\cmidrule(lr){4-9}\cmidrule(lr){10-16}\cmidrule(lr){17-18}\cmidrule(lr){19-21}
&  \rotatebox{90}{\textbf{WD1-test}} & \rotatebox{90}{\textbf{WD2}} & \rotatebox{90}{\textbf{Sweetness}} &  \rotatebox{90}{\textbf{Saltiness}} &  \rotatebox{90}{\textbf{Sourness}} &  \rotatebox{90}{\textbf{Bitterness}} &  \rotatebox{90}{\textbf{Umaminess}} &  \rotatebox{90} {\textbf{Fattiness}} &\rotatebox{90}{\textbf{Lightness}} &  \rotatebox{90}{\textbf{Grain size}} &  \rotatebox{90}{\textbf{Roughness}} &  \rotatebox{90}{\textbf{Shininess}} &  \rotatebox{90}{\textbf{Organisation}}  &  \rotatebox{90}{\textbf{Variability}} &  \rotatebox{90}{\textbf{Density}} &  \rotatebox{90}{\textbf{Movies}} &  \rotatebox{90}{\textbf{Books}} &  \rotatebox{90}{\textbf{Size}} & \rotatebox{90}{\textbf{Height}} & \rotatebox{90}{\textbf{Mass}} & \rotatebox{90}{\textbf{Average}} \\
\midrule
\multicolumn{22}{c}{\textsc{Pointwise}}\\
\midrule
Llama2-7B	&	80.5	&	61.0	&	62.8	&	53.2	&	47.2	&	52.6	&	58.2	&	65.0	&	62.0	&	60.2	&	56.4	&	42.0	&	53.6	&	61.2	&	72.2	&	59.3	&	52.8	&	68.0 &	70.0 &	50.0 &	59.4	\\
Llama2-13B	&	79.8	&	58.7	&	52.8	&	70.4	&	51.2	&	52.8	&	65.2	&	67.2	&	66.4	&	49.6	&	43.2	&	52.6	&	57.0	&	57.2	&	65.2	&	60.9	&	55.0	&	69.6 &	76.4 &	58.4 &	60.5\\
Llama3-8B & 78.6 &	59.9 & 63.8 & 74.6 & 58.2 & 58.0 & 63.8 & 64.4 & 59.0 & 58.0 & 45.2 & 53.2 & 57.4 & 53.2 & 64.4 & 65.7 & 56.5 & 83.2 & 70.2 & 61.6 & 62.5\\
Mistral-7B	&	78.3	&	61.4	&	70.2	&	69.4	&	\textbf{64.8}	&	\textbf{59.2}	&	67.8	&	68.8	&	61.0	&	57.4	&	42.4	&	47.8	&	\textbf{61.0}	&	52.4	&	56.0	&	62.4	&	59.3	&	85.6 &	70.0 &	61.0 &	62.8	\\
\midrule
\multicolumn{22}{c}{\textsc{Pairwise}}\\
\midrule
Llama2-7B	&	81.8	&	61.6	&	59.0	&	59.8	&	52.0	&	53.8	&	60.8	&	61.8	&	50.8	&	62.6	&	52.2	&	46.6	&	56.0	&	55.8	&	64.4	&	57.0	&	60.1	&	86.2 &	81.2 &	68.0 &	61.6	\\
Llama2-13B	&	82.8	&	68.0	&	58.6	&	67.4	&	50.8	&	53.6	&	67.6	&	67.6	&	50.2	&	66.8	&	58.4	&	52.0	&	55.8	&	58.8	&	68.8	&	58.3	&	55.6	&	93.8 &	91.2 &	66.2 &	64.6	\\
Llama3-8B & \textbf{83.5} & 66.2 & 68.2 & 79.8 & 61.2 & 56.8 & \textbf{72.2} & 69.2 & 58.6 & \textbf{75.0} & \textbf{60.6} & 53.0 & 54.6 &	58.8 &	73.6 &	62.5 &	56.7 &	94.0 &	94.2 &	63.2 &	68.1\\
Mistral-7B	&	82.2	&	64.2	&	59.4	&	69.0	&	52.4	&	52.4	&	66.8	&	63.0	&	58.6	&	55.0	&	52.6	&	47.8	&	54.8	&	52.0	&	58.8	&	53.3	&	52.3	&	92.6 &	88.0 &	\textbf{68.2} &	62.2	\\
\midrule
\multicolumn{22}{c}{\textsc{Baselines}}\\
\midrule
ChatGPT & 55.3 & 60.9 & 60.4 & 58.4 & 52.4 & 51.0 & 53.2 & 54.2 & 60.4 & 60.2 & 57.0 & 51.4 & 53.2 & 55.2 & 62.8 & 63.8 & \textbf{67.2} & 77.8 & 70.8 & 58.6 &	59.2\\
GPT-4 & 77.2 & \textbf{78.3} & \textbf{76.6} & \textbf{80.6} & 62.6 & 56.2 & 69.2 & \textbf{73.8} & \textbf{72.8} & 70.2 & 56.6 & \textbf{62.4} & 59.4 & \textbf{63.6} & \textbf{74.0} & \textbf{67.4} & 66.9 & \textbf{99.2} & \textbf{95.2} & 64.0 &	\textbf{71.3}\\
\bottomrule
\end{tabular}
\caption{Comparison of different models in terms of accuracy ($\%$), when classifying pairwise judgments. The pointwise and pairwise models are trained on the training split of WD1.\label{tabMainComparisonModels}}
\end{table*}

%*********************************
\paragraph{Taste}
Following \citet{chatterjee-etal-2023-cabbage}, we use a dataset with ratings about the taste of 590 food items along six dimensions: sweetness, sourness, saltiness, bitterness, fattiness and umaminess. The dataset was created by \citet{martin2014creation}, who used a panel of twelve experienced food assessors to rate the items. We use the version of the dataset that was cleaned by \citet{chatterjee-etal-2023-cabbage}, who altered some of the descriptions of the items to make them sound more natural in prompts.\footnote{Available from \url{https://github.com/ExperimentsLLM/EMNLP2023_PotentialOfLLM_LearningConceptualSpace}.}

%*********************************
\paragraph{Rocks}
\citet{nosofsky2018toward} created a dataset of rocks, with the aim of studying how cognitively meaningful representation spaces for complex domains can be learned. A total of 30 rock types were studied (10 igneous rocks, 10 metamorphic rocks and 10 sedimentary rocks). For each type of rock, 12 pictures were obtained, and each picture was annotated along 18 dimensions. However, only 7 of the considered dimensions allow for ranking all types: lightness of colour, average grain size, roughness, shininess, organisation, variability of colour and density. For our experiments, we only considered these dimensions. The dataset from \citet{nosofsky2018toward} contains ratings for each of the 12 pictures of a given rock type, where each picture was assessed by 20 annotators. To construct rankings of rock types, we average the ratings across the 12 pictures. As such, we end up with 7 rankings of 30 rock types.

%*********************************
\paragraph{Tag Genome}
\citet{DBLP:journals/tiis/VigSR12} collected a dataset\footnote{Available from \url{https://grouplens.org/datasets/movielens/tag-genome-2021/}.} of \textbf{movies}, called the Tag Genome, by asking annotators to what extent different tags apply to different movies, on a scale from 1 to 5. From these tags, we first selected those that correspond to adjectives and for which ratings for at least 15 movies were available. We then manually identified 38 of these adjectives which correspond to ordinal features.  More recently, \citet{DBLP:conf/chiir/KotkovMMSNG22} created a similar dataset for \textbf{books}. We again selected adjectives for which at least 15 items were ranked, and manually identified 32 adjectives that correspond to ordinal features. A list of the adjectives that we considered is provided in the appendix. It should be noted that most items are only judged by a single annotator, and the judgements were moreover obtained using crowdsourcing. The movies and books datasets are thus clearly noisier than the taste and rocks datasets. For this reason, we will only consider aggregated results across all tags when evaluating on these datasets. We will write \textbf{TG} to refer to the combined dataset, containing both the books and movies rankings.

%*********************************
\paragraph{Physical Properties}
Following \citet{li-etal-2023-language-models}, we consider three physical properties: mass, size and height. The ground truth for \textbf{mass} dataset was obtained from a dataset about household objects from \citet{DBLP:conf/corl/StandleySCS17}. Following \citet{chatterjee-etal-2023-cabbage}, we removed 7 items, because their mass cannot be assessed without the associated image: \emph{big elephant}, \emph{small elephant}, \emph{Ivan's phone}, \emph{Ollie the monkey}, \emph{Marshy the elephant}, \emph{boy doll} and \emph{Dali Clock}. The resulting dataset has 49 items. 
For \textbf{size} and \textbf{height}, we use the datasets from \citet{liu-etal-2022-things} as ground truth. These datasets each consist of 500 pairwise judgements.

\begin{table*}[t]
\footnotesize
\centering
\fontsize{7.7pt}{9.24pt}\selectfont
\setlength\tabcolsep{3pt}
\begin{tabular}{l cc cccccc ccccccc cc ccc}
\toprule
& \multicolumn{2}{c}{\textbf{Wikidata}} & \multicolumn{6}{c}{\textbf{Taste}} & \multicolumn{7}{c}{\textbf{Rocks}} & \multicolumn{2}{c}{\textbf{TG}} & \multicolumn{3}{c}{\textbf{Phys}}\\
\cmidrule(lr){2-3}\cmidrule(lr){4-9}\cmidrule(lr){10-16}\cmidrule(lr){17-18}\cmidrule(lr){19-21}
&  \rotatebox{90}{\textbf{WD1-test}} & \rotatebox{90}{\textbf{WD2}} & \rotatebox{90}{\textbf{Sweetness}} &  \rotatebox{90}{\textbf{Saltiness}} &  \rotatebox{90}{\textbf{Sourness}} &  \rotatebox{90}{\textbf{Bitterness}} &  \rotatebox{90}{\textbf{Umaminess}} &  \rotatebox{90} {\textbf{Fattiness}} &\rotatebox{90}{\textbf{Lightness}} &  \rotatebox{90}{\textbf{Grain size}} &  \rotatebox{90}{\textbf{Roughness}} &  \rotatebox{90}{\textbf{Shininess}} &  \rotatebox{90}{\textbf{Organisation}} &  \rotatebox{90}{\textbf{Variability}} &  \rotatebox{90}{\textbf{Density}} & \rotatebox{90}{\textbf{Movies}} & \rotatebox{90}{\textbf{Books}} &  \rotatebox{90}{\textbf{Size}} &  \rotatebox{90}{\textbf{Height}} &  \rotatebox{90}{\textbf{Mass}} \\
\midrule
Few-shot	& 50.8 & 53.6 &	52.2 & 51.8 & 51.2 & 49.2 &	47.8 & 53.4 & 55.2 & 51.4 & 45.2 & 50.8 & 49.2 &	48.6 & 51.4 & 51.5 & 54.9 &	58.8 & 54.2 & 49.6 \\
WD1-train	&	\textbf{83.5} &	66.2 &	68.2 & 79.8	& 61.2 & 56.8 & 72.2 & 69.2 & 58.6 & 75.0 & 60.6 & 53.0 & 54.6 & 58.8 & 73.6 & 62.5 & 56.7 & \textbf{94.0} & \textbf{94.2} & \textbf{63.2} \\
WD	&	-	&	-	& 63.6 & 75.2 & 57.2 & 55.2 & 69.0 & 66.8 & 68.4 & 71.4 & 63.2 & 56.2 & 57.2 & \textbf{65.6} & 73.0 & \textbf{69.0} & \textbf{60.3} & 92.8 & 93.6 & 61.6 \\
TG	&	68.3 &	\textbf{68.2} & \textbf{75.6} & 80.0 & 62.8 & \textbf{58.8} & 71.4 & 69.8 & \textbf{70.6} & 73.0 & 58.2 & 62.0 & 58.6 &	58.8 & \textbf{73.2} & - & - & 84.4 & 83.2 & 58.4 \\
Taste	&	64.0 & 64.3 & - & - & - & - & - & - & 63.2 & 77.2 & 61.6 & \textbf{65.0} & 54.0 & 58.4 & 72.6 & 65.2 & 56.0 & 77.4 & 87.8 & 51.2 \\
WD+TG+Taste	& - & - & - & - & - & - & - & - & 69.2 & \textbf{80.0} & \textbf{64.4} & 59.8 & \textbf{59.4} & \textbf{65.6} & 73.0 & - & - & 84.0 & 91.0 & 55.6 \\
WD+TG+Rocks	& - & - & 74.0 & \textbf{81.6} & \textbf{66.2} & 57.8 & \textbf{73.2} & \textbf{70.4} & - & - & - & - & - & - & - & - & - & 91.2 & 89.4 & 56.8 \\
WD+Taste+Rocks	&	-	&	-	&	-	&	-	&	-	&	-	&	-	&	-	&	-	&	-	&	-	&	-	&	-	&	-	&	-	& 68.5 & 60.1 & 85.2 & 91.8 & 56.0 \\
\bottomrule
\end{tabular}
\caption{Comparison of different models in terms of accuracy ($\%$), when classifying pairwise judgments. All results are for the pairwise model with Llama3-8B.\label{tabMainComparisonTrainingData}}
\end{table*}

\begin{table}[t]
\footnotesize
\centering
\setlength\tabcolsep{3pt}
\begin{tabular}{l cccccc }
\toprule
&  \rotatebox{90}{\textbf{Sweetness}} &  \rotatebox{90}{\textbf{Saltiness}} &  \rotatebox{90}{\textbf{Sourness}} &  \rotatebox{90}{\textbf{Bitterness}} &  \rotatebox{90}{\textbf{Umaminess}} &  \rotatebox{90} {\textbf{Fattiness}} \\
\midrule
Pointwise & 59.3 & 59.9 & 24.0 & 32.1 & 44.1 & 60.4\\
SVM (5 samples) & 60.4 & 65.9 & 48.0 & 38.3 & 59.6 & 62.3\\
SVM (30 samples) & \textbf{66.4} & \textbf{73.5} & \textbf{50.9} & \textbf{45.4} & \textbf{63.9} & \textbf{67.2}\\
Count (5 samples) & 59.9 & 66.5 & 47.3 & 41.1 &	59.5 & 60.5\\
%Count (16 pairs) & xx.x & xx.x & xx.x & xx.x & xx.x & x.xx\\
Count (30 samples) & 65.3 & 72.2 & 48.8 & 44.9 & 63.0 & 66.8\\
%\usashi{Count (100 pairs)} & 0.17 & xx.x & xx.x & xx.x & xx.x & xx.x\\
%\usashi{Greedy (32 pairs)} & 0.15 & xx.x & xx.x & xx.x & xx.x & xx.x\\
\midrule
Ada$^*$ & 17.5 & 8.5 & 12.2 & 16.4 & 22.5 & 10.7  \\
Babbage$^*$ & 19.5 & 51.1 & 20.2 & 22.0 & 22.6 & 16.0  \\
Curie$^*$ & 36.0 & 46.3 & 32.8 & 23.2 & 22.6 & 31.7  \\
Davinci$^*$ & 55.0 & 63.2 & 33.3 & 27.2 & 57.0 & 52.0  \\
\bottomrule
\end{tabular}
\caption{Comparison of ranking strategies, in terms of Spearman $\rho$\%. Baseline results marked with $^*$ were taken from \citet{chatterjee-etal-2023-cabbage}. All other results are obtained with Llama3-8B trained on WD+TG+Rocks.\label{tabRankingExperiment}}
\end{table}

\begin{table*}
\centering
%\footnotesize
\fontsize{7.7pt}{9.24pt}\selectfont
\begin{tabular}{lp{185pt}p{199pt}}
\toprule
\textbf{Feature} & \textbf{Top ranked entities} & \textbf{Bottom ranked entities}\\
\midrule
Sweetness & caramel custard, sweet cookies with chocolate, sweet pancake with sugar, syrup with water, macaroons, white chocolate, ice cream, litchi juice, ice cream bars, fruit candy
 &hake in tinfoil, unsalted pasta, tripe, raw horseradish with salt, mixed salad starch based with cold cuts, gizzards, roast beef, dry sausage, pasta with salt, burbot
\\
\midrule
Saltiness & dry-cured ham, parmesan cheese, canned sardines, canned anchovies, cooked ham, salty crackers, fried anchovies, cooked ham with salted butter, kidneys with gravy, gorgonzola cheese
& strawberry with sugar, pear, grape juice, pineapple juice, fruit crumble, strawberry juice, strawberry, banana, apple juice, blueberry\\
\midrule
Scary & Shining, The (1980), Silence of the Lambs, The (1991), Exorcist, The (1973), Psycho (1960), Ring, The (2002), Grudge, The (2004), Amityville Horror, The (2005), Texas Chainsaw Massacre, The (1974), Descent, The (2005), American Werewolf in London, An (1981) & Stand by Me (1986), Kung Fu Panda (2008), Super Size Me (2004), Station Agent, The (2003), Miss Congeniality (2000), School of Rock (2003), Roger and Me (1989), Jerry Maguire (1996), Ninotchka (1939), Driving Miss Daisy (1989)\\
\midrule
Funny & Blazing Saddles (1974), This Is Spinal Tap (1984), Monty Python's The Meaning of Life (1983), Young Frankenstein (1974), Monty Python and the Holy Grail (1975), Fish Called Wanda, A (1988), Jackass Number Two (2006), Some Like It Hot (1959), Ferris Bueller's Day Off (1986), There's Something About Mary (1998)
& Night of the Living Dead (1968), Battleship Potemkin (1925), House of Flying Daggers (Shi mian mai fu) (2004), Jesus Camp (2006), War of the Worlds (2005), Fire in the Sky (1993), Pretty Baby (1978), Ring, The (2002), Titanic (1997), United 93 (2006)\\
\midrule
Population & People's Republic of China, India, Pakistan, Egypt, Nigeria, Russia, United States of America, Brazil, Mexico, Turkey
& Dominica, Niue, Tuvalu, Antigua and Barbuda, Cook Islands, Sint Maarten, Nauru, Liechtenstein, Republic of Artsakh, Andorra\\
\bottomrule
\end{tabular}
\caption{We show the top and bottom ranked entities for five features: \emph{sweetness} and \emph{saltiness} from the \textit{food} dataset, \emph{scary} and \emph{funny} from movies, and \emph{countries population} from WD2. Results were obtained with the pairwise Llama3-8B model trained on WD+TG+Rocks for \emph{sweetness} and \emph{saltiness}, trained on WD+Taste+Rocks for \emph{scary} and \emph{funny}, and trained on WD1 for \emph{population}. \label{tabLLaMA-3_Qualitative}}
\end{table*}

%****************************************************
\section{Experiments}
We now evaluate the performance of the fine-tuning strategies on the considered datasets.\footnote{Our datasets, code and pre-trained models will be shared upon acceptance.}

\paragraph{Comparing Models} 
Table \ref{tabMainComparisonModels} compares a number of different models. We test four different LLMs: the 7B and 13B parameter Llama 2 models, the 8B parameter Llama 3 model\footnote{We use the \texttt{llama-2-7b-hf}, \texttt{llama-2-13b-hf}, and \texttt{meta-llama/Meta-Llama-3-8B} models available from \url{https://huggingface.co/meta-llama}.}, and the 7B parameter Mistral model\footnote{We use the \texttt{mistral-7b-v0.1} model available from \url{https://huggingface.co/mistralai/Mistral-7B-v0.1}.}. 
%For Llama 2 and Mistral, we use the default prompt, as explained in Sections \ref{secPairwise} and \ref{secPointwise}. For CodeLlama, we have used the following, more structured prompt:
%\begin{myquote}
%\todo{Add prompt for CodeLlama here.}
%\end{myquote}
%For instance, \todo{example of the prompt for the comparison of River Thames and Seine in terms of length}. 
We evaluate the different models in terms of their accuracy on pairwise judgements. To this end, for a given dataset, we randomly sample pairs of entities $e_i,e_j$ and construct queries asking whether $f(e_i)>f(e_j)$. For \textit{WD}, \textit{Taste} and \textit{Rocks}, we sample 500 such pairs for each of the features. Since the \textit{TG} dataset has a total of 70 features, we limit the test set to 100 pairs per feature. For this analysis, we have split the \textit{WD1} dataset into two parts: 80\% of the entities, for each feature, are used for training the models. The remaining 20\% are used as a test set. All models are fine-tuned on the training split of \textit{WD1} (apart from the baselines, which are evaluated zero-shot). This allows us to see how well the models perform on the features they were trained on (for \textit{WD1-test}) and how they generalise to unseen properties.

The aim of the analysis in Table \ref{tabMainComparisonModels} is to assess whether models can be successfully fine-tuned using a relatively small training set (i.e.\ \textit{WD1-train}), involving only well-defined numerical features. In particular, we want to test whether models which are fine-tuned on such features would also generalise to more subjective and less readily available ones, similar to the easy-to-hard generalisation that has been observed for LLMs in other tasks \cite{DBLP:journals/corr/abs-2401-06751}. The results show that this is indeed the case, at least to some extent. Overall, we can see that Mistral-7B achieves the best results among the pointwise models, while Llama3-8B achieves the best results among the pairwise models. 
The pointwise Mistral-7B model outperforms the pairwise Mistral-7B model, which is surprising given that pairwise models generally perform better in ranking tasks. The performance of the models across different features is not always consistent. Each model achieves close to random chance on some of the features, but the features where one model performs poorly are not always the same features where other models perform poorly. However, for \emph{bitterness} and \emph{roughness}, the accuracy of all models is below 61\%.
%all models perform below 60\% F1. 
Furthermore, \emph{sourness} and \emph{organisation} also stand out as being more challenging. Regarding the baselines, GPT-4 generally performs better than the fine-tuned models. Nonetheless, the pairwise Llama3-8B and pointwise Mistral-7B models outperform GPT-4 in several cases. ChatGPT performs worse on most features, but achieves the best results for \emph{books}.

\paragraph{Comparing Training Sets} 
The fine-tuned models in Table \ref{tabMainComparisonModels} generally underperform GPT-4. This may be partially explained by the fact that a small training set was used, which moreover only covered numerical features. In Table \ref{tabMainComparisonTrainingData}, we evaluate the impact of using different training sets.  For this analysis, we use the pairwise Llama3-8B model, the best fine-tuned model in Table \ref{tabMainComparisonModels}. Our focus is on seeing whether models trained on one domain can generalise to other domains. To put the results in context, we also compare with a model that is not fine-tuned but includes three in-context demonstrations (shown as \emph{few-shot}); see Appendix \ref{Appendix: Futher Details} for details. The results are again evaluated in terms of accuracy, using the same pairwise judgements as for Table \ref{tabMainComparisonModels}. \textit{WD} refers to the full dataset (including both the training and test splits of \textit{WD-1}). 

We can see in Table \ref{tabMainComparisonTrainingData} that training on larger datasets indeed leads to considerably better results. While this is not unexpected, we can also make more striking observations. For instance, the model that was trained on \textit{WD+TG+Taste} achieves strong results on \textit{Rocks}, despite covering very different features. Similarly, the model that was trained on \textit{TG} alone achieves strong results for both \textit{Taste} and \textit{Rocks}. This suggests that the fine-tuned models are indeed capable of generalising to unseen domains. However, to achieve strong results, it appears to be important that the training data contains subjective or perceptual features. In particular, training on \textit{WD} alone leads to clearly worse results on \textit{Taste} and \textit{Rocks}.
The best results in Table \ref{tabMainComparisonTrainingData} generally outperform the GPT-4 results from Table \ref{tabMainComparisonModels}. As the training and test sets cover disjoint domains, the results reflect the knowledge that is captured by the LLMs themselves, rather than knowledge that was injected during the fine-tuning process. This suggests that pre-trained LLMs capture more perceptual knowledge than it may initially appear.

\begin{table*}[t]
\centering
\footnotesize
\begin{tabular}{p{130pt}p{290pt}l}
\toprule
\textbf{Feature} & \textbf{Ranking} \\
\midrule
Is $X$ more suitable as a present for a 10-year-old girl than $Y$?
&1.\ lego, 2.\ bicycle, 3.\ lipgloss, 4.\ puppy, 5.\ airpods, 6.\ guitar, 7.\ helicopter, 8.\ Famous Five Collection - Enid Blyton, 9.\ broom, 10.\ watch, 11.\ gold fish, 12.\ salmon fish, 13.\ knife, 14.\ crocodile, 15.\ coffee mug, 16.\ Sidney Sheldon Series\\
\midrule
Does $X$ require more urgent medical treatment than $Y$? 
& 1.\ cardiac arrest, 2.\ brain haemmorage, 3.\ stroke, 4.\ infection, 5.\ flu, 6.\ cataract surgery, 7.\ cough and cold, 8.\ dental fillings, 9.\ fist bump, 10.\ paper cut\\
\midrule
Is $X$ more important than $Y$ post marathon? 
&1.\ run marathon, 2.\ hydrate, 3.\ indulge in alcohol, 4.\ partying, 5.\ shopping, 6.\ rest, 7.\ skip meal, 8.\ snacking, 9.\ haircut
\\
\midrule
Is $X$ cheaper to live than $Y$? 
& 1.\ Mexico City, 2.\ Budapest, 3.\ Bali, 4.\ Bangalore, 5.\ Bangkok, 6.\ Paris, 7.\ Singapore, 8.\ New York, 9.\ Los angeles, 10.\ Tokyo\\
\midrule
Does $X$ burn more calories than $Y$?  
&1.\ running, 2.\ swimming, 3.\ stair climbing, 4.\ eating, 5.\ strolling, 6.\ tai chi, 7.\ pilates, 8.\ drinking beer, 9.\ sleeping, 10.\ watching netflix
\\
\midrule
Does the salary of $X$ exceed that of $Y$?  
&1.\ entrepreneur, 2.\ doctor, 3.\ lawyer, 4.\ consultant, 5.\ engineer, 6.\ scientist, 7.\ lecturer, 8.\ soldier, 9.\ mechanic, 10.\ dance teacher
\\
\midrule
Is $X$ more suitable for long-distance travel with a family than $Y$?
&1.\ spaceship, 2.\ minivan, 3.\ aeroplane, 4.\ freight train, 5.\ local train, 6.\ speed boat, 7.\ submarine, 8.\ taxi, 9.\ quad bike, 10.\ motor cycle
\\
\midrule
Is $X$ more likely than $Y$ to be the first thing you do in the morning?
&1.\ hygiene routine, 2.\ hitting disco, 3.\ watch movies, 4.\ making bed, 5.\ socialize, 6.\ getting ready for shopping, 7.\ attend in-person class, 8.\ have snack, 9.\ checking emails, 10.\ karaoke
\\
\midrule
Is $X$ more important than $Y$ when selecting a primary school for children?
&1.\ alumni network, 2.\ parent's income, 3.\ academic reputation, 4.\ nearby shopping centre, 5.\ swimming pool facilities, 6.\ breakfast club availability, 7.\ playground, 8.\ classroom size, 9.\ parking facilities, 10.\ school location
\\
\midrule
Is $X$ more urgent than $Y$ when considering home improvement projects?
&1.\ bathroom renovation, 2.\ repairing broken windows, 3.\ fixing a leak, 4.\ installing new locks, 5.\ landscaping the backyard, 6.\ repairing door lock, 7.\ installing new furniture, 8.\ installing new light, 9.\ painting ceilings, 10.\ changing carpet
\\
\bottomrule
\end{tabular}
\caption{Examples of entity rankings for selected (commonsense) features. Results were obtained with the pairwise Llama3-8B model trained on WD+TG+Rocks. \label{tabCommonsenseRanking}}
\end{table*}

\paragraph{Comparing Ranking Strategies} 
Table \ref{tabRankingExperiment} compares different strategies for generating rankings. The pointwise model can be directly used for this purpose. For the pairwise model, we show results with the SVM strategy and the Count strategy. We furthermore vary the number of pairwise judgments per entity (5 or 30) for Count and SVM.  For this experiment, we use the pointwise and pairwise Llama3-8B models that were trained on WD+TG+Rocks. We evaluate the different models by comparing the predicted rankings with the ground truth in terms of Spearman $\rho$. We can see that the pairwise approaches consistently outperform the pointwise model. %This is somewhat surprising, given the strong performance of the pointwise models in Table \ref{tabMainComparisonModels}. Essentially, because the ranking strategies aggregate many pairwise samples, the noisy nature of the pairwise judgments can to some extent be mitigated. 
The SVM method generally performs better than the Count method, especially in the case where 30 judgments per entity are obtained. We also compare with the GPT-3 results reported by \citet{chatterjee-etal-2023-cabbage}, finding that the pairwise Llama model consistently performs best.

%\begin{table}[t]
%\footnotesize
%\centering
%\setlength\tabcolsep{3pt}
%\begin{tabular}{l llllll }
%\toprule
%&  \rotatebox{90}{\textbf{Sweetness}} &  \rotatebox{90}{\textbf{Saltiness}} &  \rotatebox{90}{\textbf{Sourness}} &  \rotatebox{90}{\textbf{Bitterness}} &  \rotatebox{90}{\textbf{Umaminess}} &  \rotatebox{90} {\textbf{Fattiness}} \\
%\midrule
%Prompt 1 & xx.x & xx.x & xx.x & xx.x & xx.x & xx.x\\
%Prompt 2 & xx.x & xx.x & xx.x & xx.x & xx.x & xx.x\\
%Prompt 3 & xx.x & xx.x & xx.x & xx.x & xx.x & xx.x\\
%Prompt 4 & xx.x & xx.x & xx.x & xx.x & xx.x & xx.x\\
%Prompt 5 & xx.x & xx.x & xx.x & xx.x & xx.x & xx.x\\
%\bottomrule
%\end{tabular}
%\caption{Comparison of different prompts, in terms of accuracy ($\%$), when classifying pairwise judgments. All results are for the pairwise Llama2-13B model, fine-tuned on WD+TG+Rocks. \label{tabPromptComparison}}
%\end{table}

\paragraph{Qualitative Analysis}
Table \ref{tabLLaMA-3_Qualitative} shows the 10 highest and lowest ranked entities for some selected features, according to the rankings from the SVM method with the pairwise Llama3-8B model. The results for \textit{sweetness} and \textit{saltiness} were obtained with the model that was trained on WD+TG+Rocks. The rankings for \textit{scary} and \textit{funny} movies were obtained with the model that was trained on WD+Taste+Rocks. The ranking for \textit{population} was obtained with the model that was trained on WD1. The table shows that the model was successful in selecting these top and bottom ranked entities. The top-ranked entities for \textit{sweetness}, for instance, are all clearly sweet food items, while none of the bottom ranked entities are. Similar observations can be made for the other features. The model is sometimes less successful in distinguishing middle-ranked entities from bottom-ranked entities. For instance, most cheeses appear at the bottom of the ground truth ranking, whereas the model predicted these to be somewhere closer to the middle.\footnote{A more detailed analysis  can be found in the appendix.}  For \emph{population}, we can see that while the top-ranked entities are all countries with a high population, their relative ranking is not accurate. 
%For instance, Egypt is ranked in fourth place, despite not appearing in the top-10 of the most populous countries.

To further test the ability of the model to generalise to unseen properties, Table \ref{tabCommonsenseRanking} shows results that were obtained for a number of selected features, focusing primarily on commonsense properties. For instance, for the first example, we compared a number of items according to their suitability as a present for a 10-year-old girl. To obtain the rankings, we exhaustively compared every pair of entities (from the considered set) and used the SVM method. The results were obtained with the pairwise Llama3-8B model that was trained on WD+TG+Rocks. These examples illustrate that the model often struggles with commonsense features. For instance, \emph{spaceship} is listed as the most suitable vehicle for long-distance travel with a family, the \emph{school location} is considered to be the least important criterion for selecting a primary school and a \emph{bathroom renovation} is considered to be more urgent than \emph{fixing a leak}. %, and \emph{indulge in alcohol} is considered more important than \emph{snacking} post marathon.
%a \emph{helicopter} is considered to be a more suitable present for a 10-year-old girl than a \emph{coffee mug}. 
The disappointing results for commonsense features are somewhat surprising, although \citet{chatterjee2023probing} similarly found that ChatGPT struggled with commonsense properties.

%****************************************************
\section{Conclusions}
We have studied the problem of ranking entities along conceptual space dimensions, such as sweetness (for food), roughness (for rocks) or scary (for movies). We found that fine-tuning LLMs on data from one domain (e.g.\ taste) is a viable strategy for learning to extract rankings in unrelated domains (e.g.\ rocks), as long as both domains are perceptual. In contrast, LLMs that were fine-tuned on objective numerical features from Wikidata were less successful when applied to perceptual domains. When comparing pairwise and pointwise strategies, surprisingly, we found that pointwise methods were as successful as pairwise methods for making pairwise judgements (i.e.\ should entity $e_1$ be ranked before entity $e_2$), although pairwise methods still had the advantage when such judgments were aggregated. Overall, our results suggest that the current generation of open-source LLMs, such as Llama and Mistral, can be effectively used for constructing high-quality conceptual space representations. However, further work is needed to construct more comprehensive training sets. Encouragingly, we found that subjective (and relatively noisy) rankings, such as those from the movies and books datasets, can also be effective, while being much easier to obtain than  perceptual features.

\paragraph{Acknowledgments}
This work was supported by EPSRC grants EP/V025961/1 and EP/W003309/1.

%******************************************************************
\section*{Limitations}
The performance of LLMs is highly sensitive to the prompting strategy. While we have made efforts to choose a reasonable prompt, it is likely that better results are possible with different choices. Furthermore, while we have tested a number of different LLMs, it is possible that other (existing or future) models of similar sizes may behave qualitatively different. Care should therefore be taken when drawing any conclusions about the limitations of LLMs in general. Moreover, the limitations we have identified might be particular to the specific fine-tuning techniques that we have used, rather than reflecting limitations of the underlying LLMs. When it comes to modelling subjective features, such as those in the movies and books datasets, it is important to acknowledge that people may have different points of view. When using conceptual space representations extracted from LLMs in downstream applications, we thus need to be aware that these representations are biased and, at best, can only represent a majority opinion.

\bibliography{anthology,custom}

\appendix

\begin{table*}[t]
\fontsize{7.7pt}{9.24pt}\selectfont
\centering
\setlength\tabcolsep{3pt}
\begin{tabular}{l cc cccccc ccccccc cc ccc }
\toprule
& \multicolumn{2}{c}{\textbf{Wikidata}} & \multicolumn{6}{c}{\textbf{Taste}} & \multicolumn{7}{c}{\textbf{Rocks}} & \multicolumn{2}{c}{\textbf{TG}} & \multicolumn{3}{c}{\textbf{Phys}}\\
\cmidrule(lr){2-3}\cmidrule(lr){4-9}\cmidrule(lr){10-16}\cmidrule(lr){17-18}\cmidrule(lr){19-21}
&  \rotatebox{90}{\textbf{WD1-test}} & \rotatebox{90}{\textbf{WD2}} & \rotatebox{90}{\textbf{Sweetness}} &  \rotatebox{90}{\textbf{Saltiness}} &  \rotatebox{90}{\textbf{Sourness}} &  \rotatebox{90}{\textbf{Bitterness}} &  \rotatebox{90}{\textbf{Umaminess}} &  \rotatebox{90} {\textbf{Fattiness}} &\rotatebox{90}{\textbf{Lightness}} &  \rotatebox{90}{\textbf{Grain size}} &  \rotatebox{90}{\textbf{Roughness}} &  \rotatebox{90}{\textbf{Shininess}} &  \rotatebox{90}{\textbf{Organisation}}  &  \rotatebox{90}{\textbf{Variability}} &  \rotatebox{90}{\textbf{Density}} &  \rotatebox{90}{\textbf{Movies}} &  \rotatebox{90}{\textbf{Books}} &  \rotatebox{90}{\textbf{Size}} & \rotatebox{90}{\textbf{Height}} & \rotatebox{90}{\textbf{Mass}}  \\
\midrule
ChatGPT & 0.10 & 0.10 & 0.00 & 0.00 & 0.40 & 0.00 & 0.00 & 0.20 & 0.00 & 0.00 & 0.40 & 0.20 & 0.20 & 0.20 & 0.20 & 0.47 & 1.00 & 0.20 & 0.00 & 0.40\\
GPT-4 & 2.04 & 0.12 & 0.00 & 0.00 & 0.00 & 0.00 & 0.00 & 0.00 & 0.00 & 0.00 & 0.00 & 0.00 & 0.00 & 0.00 & 0.00 & 0.82 & 4.16 & 0.00 & 0.00 & 0.00\\
\bottomrule
\end{tabular}
\caption{Percentage of cases where ChatGPT and GPT-4 refused to answer a question about a pairwise comparison between two entities.\label{tabChatGPTNonResponses}}
\end{table*}

\section{Further Details}\label{Appendix: Futher Details}

\paragraph{Training Details}
%We choose Llama-2-13B \footnote{\url{https://huggingface.co/meta-llama/Llama-2-13b}} from the family of Meta models \cite{touvron2023llama} as our main model for study. The model was trained on 50000 data points considered from publicly available Wikidata resources. 
%To train the LLMs, we use a batch size of 8 and a max step size of 12500. The models were trained for 2 epochs with lora\_r = 32 and lora\_alpha = 64 respectively.

To train the four base models, 
%i.e., \textit{meta-llama/Llama-2-7b-hf}, \textit{mistralai/Mistral-7B-v0.1}, and \textit{meta-llama/Llama-2-13b-hf}, 
we used the QLoRa method, which allows converting the floating-point 32 format to smaller data types. In particular, for all three models, we used 4-bit quantization for efficient training. In the QLoRa configuration, $r$ (the rank of the low-rank matrix used in the adapters) was set to 32,  $\alpha$ (the scaling factor for the learned weights) was set to 64, and dropout was set to 0.05. We applied QLoRa to all the linear layers of the models, including \textit{q\_proj}, \textit{k\_proj}, \textit{v\_proj}, \textit{o\_proj}, \textit{gate\_proj}, \textit{up\_proj}, \textit{down\_proj}, and \textit{lm\_head}.
The models were trained with a batch size of 8. We used 20\% of the WD1 training split as a validation set. Based on this validation set, we fixed the number of training steps to 12500 for the pairwise models and 1500 for the pointwise models. Note that we need fewer training steps for the pointwise model, because each mini-batch consists of 8 entities, and we consider all pairwise combinations of these entities. In contrast, for the pairwise model, each mini-batch consists of 8 pairwise combinations. We also observed that the pointwise model converges more quickly than the pairwise model. 

%For the pairwise approaches, training was performed for 25,000 steps, while for the pointwise approaches, training was performed for 1,500 steps. This decision was based on the validation dataset.

\paragraph{OpenAI Models}
Table \ref{tabChatGPTNonResponses} shows for how many cases ChatGPT and GPT-4 failed to answer with \textit{yes} or \textit{no}, when asked about pairwise comparisons. Overall, such cases were rare. The highest number of failures was seen for TG dataset, which appears to be related to the subjective nature of the features involved.

\begin{table}
\footnotesize
\centering
\setlength\tabcolsep{4pt}
\begin{tabular}{llll}
\toprule
&\textbf{Entity type} & \textbf{Feature} & \textbf{Size}\\
\midrule
\parbox[t]{2mm}{\multirow{10}{*}{\rotatebox[origin=c]{90}{\textbf{WD1}}}}&mountain & elevation & 1000 \\
&building & height & 1000 \\
&river & length & 1000 \\
&person & \# social media followers & 1000 \\
&city & population & 1000\\
&species & mass & 1000\\
&organisation & inception date & 1000\\
&person & date of birth & 1000\\
&museum & latitude & 1000\\
&landform & area & 1000\\
\midrule
\parbox[t]{2mm}{\multirow{10}{*}{\rotatebox[origin=c]{90}{\textbf{WD2}}}} & country & population & 196{}\\
& musical object & inception date & 561{}\\
& chemical element & atomic number & 166{}\\
& chemical element & discovery date & 113{}\\
& building & \# elevators & 151{}\\
& director & \# academy awards & 65{}\\
& actor & \# academy awards & 74{}\\
& food & water footprint & 56{}\\
& composer & \# grammy awards & 71{}\\
& food & Scoville grade & 43{}\\
\bottomrule
\end{tabular}
\caption{Overview of the datasets based on Wikidata.\label{tabWikidataOverview}}
\end{table}

\begin{table}[t]
\footnotesize
\setlength\tabcolsep{4pt}
\centering
\begin{tabular}{llllll}
\toprule
\textbf{Tag} & \textbf{\#Movies} &&& \textbf{Tag} & \textbf{\#Movies}\\
\midrule
scary	&	82	&&&	grim	&	20	\\
funny	&	217	&&&	gritty	&	34	\\
gory	&	33	&&&	inspirational	&	90	\\
dark	&	139	&&&	intelligent	&	18	\\
beautiful	&	117	&&&	intense	&	53	\\
intellectual	&	32	&&&	melancholic	&	17	\\
artistic	&	91	&&&	predictable	&	121	\\
absurd	&	20	&&&	pretentious	&	29	\\
bleak	&	23	&&&	quirky	&	151	\\
bloody	&	27	&&&	realistic	&	74	\\
boring	&	186	&&&	romantic	&	46	\\
claustrophobic	&	19	&&&	sad	&	130	\\
clever	&	68	&&&	satirical	&	106	\\
complex	&	23	&&&	sentimental	&	28	\\
controversial	&	44	&&&	surreal	&	241	\\
dramatic	&	24	&&&	suspenseful	&	19	\\
emotional	&	34	&&&	tense	&	40	\\
enigmatic	&	36	&&&	violent	&	132	\\
frightening	&	18	&&&	witty	&	47	\\
\bottomrule
\end{tabular}
\caption{Considered set of tags for the Movies dataset.\label{tabMovies}}
\end{table}

\begin{table}[t]
\footnotesize
\setlength\tabcolsep{4pt}
\centering
\begin{tabular}{llllll}
\toprule
\textbf{Tag} & \textbf{\#Books} &&& \textbf{Tag} & \textbf{\#Books}\\
\midrule
absurd	&	106	&&&	literary	&	525	\\
beautiful	&	28	&&&	philosophical	&	80	\\
bizarre	&	37	&&&	political	&	138	\\
controversial	&	40	&&&	predictable	&	27	\\
cool	&	31	&&&	quirky	&	20	\\
crazy	&	23	&&&	realistic	&	36	\\
dark	&	659	&&&	romantic	&	125	\\
educational	&	121	&&&	sad	&	154	\\
funny	&	331	&&&	satirical	&	23	\\
futuristic	&	157	&&&	short	&	30	\\
gritty	&	18	&&&	silly	&	25	\\
hilarious	&	66	&&&	strange	&	22	\\
inspirational	&	195	&&&	surreal	&	28	\\
intellectual	&	17	&&&	unique	&	31	\\
intense	&	27	&&&	weird	&	46	\\
interesting	&	33	&&&	witty	&	17	\\
\bottomrule
\end{tabular}
\caption{Considered set of tags for the Books dataset.\label{tabBooks}}
\end{table}

\paragraph{Datasets}
Table \ref{tabWikidataOverview} gives an overview of the properties that were selected for the WD1 and WD2 datasets, along with the corresponding number of entities.
Table \ref{tabMovies} and \ref{tabBooks} similarly show the tags that have been considered for the Movies and Books datasets, along with the number of corresponding entities.

\paragraph{Few-shot Baseline}
For the few-shot configuration in Table \ref{tabMainComparisonTrainingData}, we use the pre-trained Llama3-8B model with the following prompt:

\begin{lstlisting}
Answer the following with Yes or No only. In the worst case, if you do not know the answer then choose randomly between Yes and No.
This question is about two rivers: Is Nile longer than Indus?
Yes
This question is about two countries: Is Japan more populated than India?
No
This question is about two countries: Is China larger than India?
Yes
\end{lstlisting}

\noindent Note that the same three in-context examples were provided for all datasets.

\begin{table*}
\centering
%\footnotesize
\fontsize{7.7pt}{9.24pt}\selectfont
\begin{tabular}{lp{185pt}p{185pt}}
\toprule
\textbf{Feature} & \textbf{Entities ranked too high} & \textbf{Entities ranked too low}\\
\midrule
Sweetness & coulommiers cheese, chaource cheese, mimolette cheese, pasta with soy sauce, latte without sugar, reblochon cheese, plain yogurt, mont d'or cheese, saint-agur cheese, faisselle & martini with lemon juice, martini, cod fritters, champignons crus with vinaigrette, hamburger, salty crackers, light lager, andouillette sausage, soft boiled eggs, omelette with vinegar\\
\midrule
Saltiness & marinated mussels, liquorice candy, fortified wines, kir, oriental pastries, cola soda, petit suisse with sugar, petit suisse with sugar and cream, aperitif with anise, petit suisse & carrot puree with cream, mix vegetables salad, moussaka, guacamole, pies, zucchini, stuffed zucchini, quiches, bulgur, broccoli with cream \\
\midrule
Scary & Pirates of the Caribbean: The Curse of the Black Pearl (2003), Interview with the Vampire: The Vampire Chronicles (1994), Scream (1996), Terminator, The (1984), Quills (2000), Batman Begins (2005), Evil Dead II (Dead by Dawn) (1987), Dawn of the Dead (1978), Spirited Away (Sen to Chihiro no kamikakushi) (2001), Requiem for a Dream (2000) & Final Fantasy: The Spirits Within (2001), Eye of the Needle (1981), Sunless (Sans Soleil) (1983), Outland (1981), Super Size Me (2004), Slumdog Millionaire (2008), Underworld (2003), Roger \& Me (1989), One Hour Photo (2002), Close Encounters of the Third Kind (1977)\\
\midrule
Funny& Fargo (1996), Original Kings of Comedy, The (2000), Meet the Spartans (2008), Simpsons Movie, The (2007), Happy Gilmore (1996), Jackass Number Two (2006), Who Framed Roger Rabbit? (1988), Tenacious D in The Pick of Destiny (2006), Men in Black (a.k.a. MIB) (1997), Elf (2003) & Ref, The (1994), American Psycho (2000), Run Lola Run (Lola rennt) (1998), Charter Trip, The (a.k.a.\ Package Tour, The) (1980), Bend It Like Beckham (2002), License to Drive (1988), Battleship Potemkin (1925), Jesus Camp (2006), Slap Shot (1977), Night of the Living Dead (1968)\\
\midrule
Population& Djibouti, Qatar, Eritrea, Botswana, Papua New Guinea, Gabon, Libya, Mongolia, Mauritania, Namibia & Burundi, Rwanda, Switzerland, Wales, Kingdom of the Netherlands, Belgium, England, Italy, Netherlands, Czech Republic\\
\bottomrule
\end{tabular}
\caption{Llama2-13B error analysis, showing the entities with the maximum difference in rank position between the ground truth ranking and the predicted ranking.
%We show the top and bottom ranked entities for five features: sweetness and saltiness, from the \textit{food} dataset, scary and funny, from movies, and countries population, from WD2.
\label{tabErrorAnalysis}}
\end{table*}

\begin{figure}
\includegraphics[width=\columnwidth]{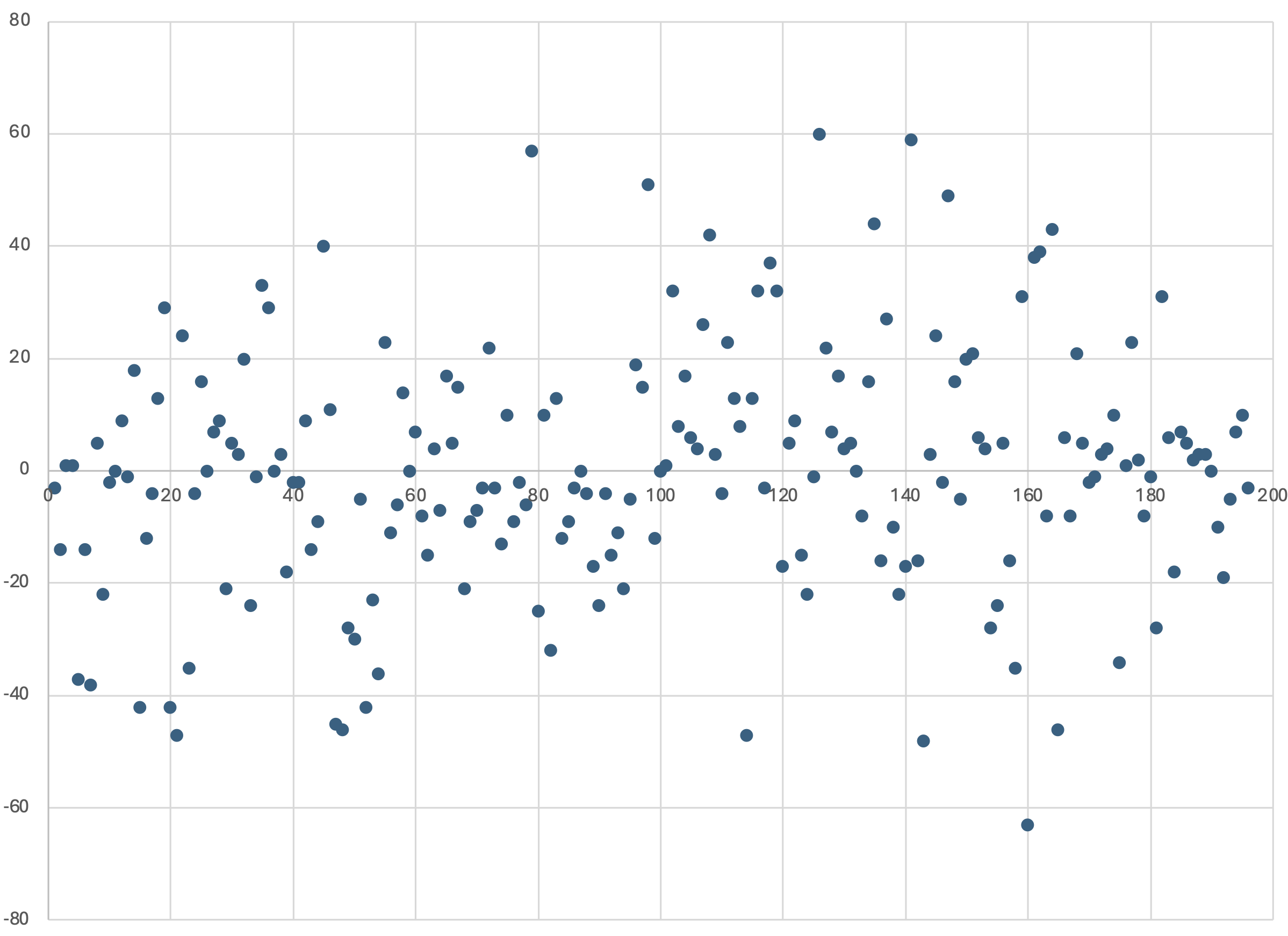}
\caption{Scatter plot comparing the popularity of Wikidata entities (X-axis) with the prediction error (Y-axis) for the \textit{countries population} feature.\label{figScatter}}
\end{figure}

%*****************************************************************
\section{Additional Analysis}

\begin{table*}[t]
\footnotesize
\centering
\fontsize{7.7pt}{9.24pt}\selectfont
\setlength\tabcolsep{3pt}
\begin{tabular}{l cc cccccc ccccccc cc ccc}
\toprule
& \multicolumn{2}{c}{\textbf{Wikidata}} & \multicolumn{6}{c}{\textbf{Taste}} & \multicolumn{7}{c}{\textbf{Rocks}} & \multicolumn{2}{c}{\textbf{TG}} & \multicolumn{3}{c}{\textbf{Phys}}\\
\cmidrule(lr){2-3}\cmidrule(lr){4-9}\cmidrule(lr){10-16}\cmidrule(lr){17-18}\cmidrule(lr){19-21}
&  \rotatebox{90}{\textbf{WD1-test}} & \rotatebox{90}{\textbf{WD2}} & \rotatebox{90}{\textbf{Sweetness}} &  \rotatebox{90}{\textbf{Saltiness}} &  \rotatebox{90}{\textbf{Sourness}} &  \rotatebox{90}{\textbf{Bitterness}} &  \rotatebox{90}{\textbf{Umaminess}} &  \rotatebox{90} {\textbf{Fattiness}} &\rotatebox{90}{\textbf{Lightness}} &  \rotatebox{90}{\textbf{Grain size}} &  \rotatebox{90}{\textbf{Roughness}} &  \rotatebox{90}{\textbf{Shininess}} &  \rotatebox{90}{\textbf{Organisation}} &  \rotatebox{90}{\textbf{Variability}} &  \rotatebox{90}{\textbf{Density}} & \rotatebox{90}{\textbf{Movies}} & \rotatebox{90}{\textbf{Books}} &  \rotatebox{90}{\textbf{Size}} &  \rotatebox{90}{\textbf{Height}} &  \rotatebox{90}{\textbf{Mass}} \\
\midrule
WD1-train	&	\textbf{82.8}	&	\textbf{68.0}	&	58.6	&	67.4	&	50.8	&	53.6	&	67.6	&	67.6	&	50.2	&	66.8	&	58.4	&	52.0	&	55.8	&	58.8	&	68.8	&	58.3	&	55.6 & \textbf{93.8} &	\textbf{91.2} &	\textbf{66.2} \\
WD	&	-	&	-	&	55.2	&	64.8	&	51.2	&	53.8	&	62.4	&	63.0	&	46.8	&	68.8	&	\textbf{60.8}	&	60.0	&	50.4	&	\textbf{64.8}	&	70.6	&	65.3	&	62.3	& 78.0 & 79.4 & 60.4 \\
TG	&	63.3	&	56.9	&	71.2	&	71.6	&	\textbf{60.0}	&	58.8	&	69.0	&	65.6	&	\textbf{71.2}	&	69.6	&	48.8	&	60.6	&	57.6	&	55.8	&	66.0	&	-	&	-	& 50.6 & 54.6 & 55.2 \\
Taste	&	62.1	&	51.1	&	-	&	-	&	-	&	-	&	-	&	-	&	66.4	&	\textbf{72.2}	&	56.8	&	\textbf{60.8}	&	58.6	&	53.2	&	74.0	&	66.2	&	55.7	& 53.0 & 61.4 & 58.2 \\
WD+TG+Taste	&	-	&	-	&	-	&	-	&	-	&	-	&	-	&	-	&	61.2	&	70.6	&	57.0	&	59.2	&	\textbf{62.8}	&	57.6	&	\textbf{78.4}	&	-	&	-	& 77.8 & 85.4 & 62.2 \\
WD+TG+Rocks	&	-	&	-	&	\textbf{74.0}	&	\textbf{72.4}	&	\textbf{60.0}	&	\textbf{60.2}	&	\textbf{70.6}	&	\textbf{72.2}	&	-	&	-	&	-	&	-	&	-	&	-	&	-	&	-	&	-	& 85.6 & 88.2 & 62.0 \\
WD+Taste+Rocks	&	-	&	-	&	-	&	-	&	-	&	-	&	-	&	-	&	-	&	-	&	-	&	-	&	-	&	-	&	-	&	\textbf{69.1}	&	\textbf{65.2} & 89.4 & \textbf{91.2} & 63.8 \\
\bottomrule
\end{tabular}
\caption{Comparison of different models in terms of accuracy ($\%$), when classifying pairwise judgments. All results are for the pairwise model with Llama2-13B.\label{tabMainComparisonTrainingDataLlama2}}
\end{table*}

\begin{table}[t]
\footnotesize
\centering
\setlength\tabcolsep{3pt}
\begin{tabular}{l cccccc }
\toprule
&  \rotatebox{90}{\textbf{Sweetness}} &  \rotatebox{90}{\textbf{Saltiness}} &  \rotatebox{90}{\textbf{Sourness}} &  \rotatebox{90}{\textbf{Bitterness}} &  \rotatebox{90}{\textbf{Umaminess}} &  \rotatebox{90} {\textbf{Fattiness}} \\
\midrule
Pointwise & 51.0 & \textbf{64.8} & 32.5 & 35.2 & 52.0 & 61.7\\
SVM (5 samples) & 62.1 & 62.2 & 42.6 & 44.6 & 56.4 & 63.0\\
SVM (30 samples) & \textbf{66.0} & 64.7 & 47.6 & 47.0 & \textbf{60.6} & \textbf{65.8}\\
Count (5 samples) & 59.0 & 57.4 & 46.7 & 41.7 & 53.5 & 60.1\\
%Count (16 pairs) & xx.x & xx.x & xx.x & xx.x & xx.x & x.xx\\
Count (30 samples) & 64.8 & 64.7 & \textbf{49.1} & \textbf{47.2} & 59.9 & 64.7\\
%\usashi{Count (100 pairs)} & 0.17 & xx.x & xx.x & xx.x & xx.x & xx.x\\
%\usashi{Greedy (32 pairs)} & 0.15 & xx.x & xx.x & xx.x & xx.x & xx.x\\
\midrule
Ada$^*$ & 17.5 & 8.5 & 12.2 & 16.4 & 22.5 & 10.7  \\
Babbage$^*$ & 19.5 & 51.1 & 20.2 & 22.0 & 22.6 & 16.0  \\
Curie$^*$ & 36.0 & 46.3 & 32.8 & 23.2 & 22.6 & 31.7  \\
Davinci$^*$ & 55.0 & 63.2 & 33.3 & 27.2 & 57.0 & 52.0  \\
\bottomrule
\end{tabular}
\caption{Comparison of ranking strategies, in terms of Spearman $\rho$\%. Baseline results marked with $^*$ were taken from \citet{chatterjee-etal-2023-cabbage}. All other results are obtained with Llama2-13B trained on WD+TG+Rocks.\label{tabRankingExperimentLlama2-13B}}
\end{table}

\paragraph{Analysis using Llama2-13B}
In the main paper, we primarily used Llama3-8B for the experiments. Tables  \ref{tabMainComparisonTrainingDataLlama2} and \ref{tabRankingExperimentLlama2-13B} show the corresponding results for Llama2-13B, focusing respectively on analysing the impact of using different training sets and on comparing different ranking strategies. Table \ref{tabLLaMA-2Qualitative} shows the 10 highest and lowest ranked entities, according to the rankings from the SVM method with the pairwise Llama2-13B model. These examples can be contrasted with the Llama3 results from Table \ref{tabLLaMA-3_Qualitative}.

\paragraph{Error Analysis} 
Table \ref{tabErrorAnalysis} presents an error analysis for the same five rankings that were considered in Table \ref{tabLLaMA-2Qualitative}. Specifically, in Table \ref{tabErrorAnalysis} we focus on the entities where the difference between the predicted ranking position and the position of the entity in the ground truth ranking is highest. On the left, we show entities which are ranked too high (i.e.\ where the model predicts the entity has the feature to a greater extent than is the case according to the ground truth). On the right, we show entities which are ranked too low. In the case of sweetness, we can see that the model consistently ranks cheeses to high. They are predicted to be in ranking positions 150-250, whereas the ground truth puts them at 500-590. In the case of saltiness, we can see that sweet drinks and pastries are ranked too high. For instance, \emph{cola soda} is ranked in position 108 whereas the ground truth puts it at 517 (out of 590). Overall, these results suggest that the model struggles with certain food groups. For the features \textit{scary}, \textit{funny} and \textit{population}, clear patterns are harder to detect.

\paragraph{Impact of Entity Popularity} The reliability of LLMs when it comes to modelling entity knowledge has been found to correlate with the popularity of the entities involved \cite{mallen-etal-2023-trust}. To analyse this aspect, Figure \ref{figScatter} compares entity popularity with prediction error, for the \textit{countries population} feature from the WD2 dataset. For this analysis, we have used the pairwise Llama2-13B model that was trained on the WD1-training split. We obtained a ranking of all countries using the SVM method with 20 samples. On the X-axis, the entities are ranked from the most popular to the least popular. On the Y-axis, we show the prediction error for the corresponding entity, measured as the difference between the position of the entity in the predicted ranking and its position in the ground truth ranking. Based on this analysis, no clear correlation between entity popularity and prediction error can be observed.

\paragraph{Accuracy Variability}
Table \ref{Standard Deviation} shows standard deviations across four runs for one of the best-performing models. As can be seen, the standard deviations are relatively small, indicating that the main differences across the models are not due to random variations.

\begin{table*}[t]
\fontsize{7.7pt}{9.24pt}\selectfont
\centering
\setlength\tabcolsep{3pt}
\begin{tabular}{l cccccc ccc }
\toprule
& \multicolumn{6}{c}{\textbf{Taste}} & \multicolumn{3}{c}{\textbf{Phys}}\\
\cmidrule(lr){2-7}\cmidrule(lr){8-10}
& \rotatebox{90}{\textbf{Sweetness}} &  \rotatebox{90}{\textbf{Saltiness}} &  \rotatebox{90}{\textbf{Sourness}} &  \rotatebox{90}{\textbf{Bitterness}} &  \rotatebox{90}{\textbf{Umaminess}} &  \rotatebox{90}{\textbf{Fattiness}} & \rotatebox{90}{\textbf{Size}} & \rotatebox{90}{\textbf{Height}} & \rotatebox{90}{\textbf{Mass}}  \\
\midrule
%WD+TG+Rocks (One Run) & 74.0 & 81.6 & 66.2 & 57.8 & 73.2 & 70.4 & 91.2 & 89.4 & 56.8\\
WD+TG+Rocks  & 73.8 $\pm$ 0.6 & 81.8 $\pm$ 0.6 & 63.7 $\pm$ 2.5 & 57.9 $\pm$ 0.6 & 72.9 $\pm$ 0.3 & 70.0 $\pm$ 0.4 & 89.9 $\pm$ 0.9 & 89.2 $\pm$ 0.8 & 58.2 $\pm$ 1.4\\
\bottomrule
\end{tabular}
\caption{Mean accuracy and standard deviations across four runs for the pairwise Llama3-8B model.\label{Standard Deviation}}
\end{table*}

\begin{table*}
\centering
%\footnotesize
\fontsize{7.7pt}{9.24pt}\selectfont
\begin{tabular}{lp{185pt}p{185pt}}
\toprule
\textbf{Feature} & \textbf{Top ranked entities} & \textbf{Bottom ranked entities}\\
\midrule
Sweetness & mango, dried date, white chocolate, peach , pineapple in syrup, fruit candy, syrup with water, ice cream, strawberry, sweet pancake with maple syrup & minced beef patty, grilled calf livers, squid, sandwich with cold cuts, gizzards, croque-monsieur, roast rabbit, stir-fried bacon, roast beef , calf head with vinaigrette\\
\midrule
Saltiness & green olives, extruded salty crackers, soy sprouts with soy sauce, canned anchovies, canned sardines, pasta with soy sauce, salted pies, marinated mussels, potato chips, salted cake & clafoutis, raspberry cake, stewed apple, raspberry with whipped cream, white chocolate, strawberry with cream and sugar, mix fruits juice, apple, raspberry, strawberry \\
\midrule
Scary & Descent, The (2005), Grudge, The (2004), Exorcist, The (1973), Silence of the Lambs, The (1991), Ring, The (2002), Texas Chainsaw Massacre, The (1974),     Shining, The (1980), Seven (a.k.a. Se7en) (1995), Amityville Horror, The (2005), American Werewolf in London, An (1981) & Super Size Me (2004), Station Agent, The (2003), Ray (2004), Dances with Wolves (1990), Jerry Maguire (1996), Driving Miss Daisy (1989), School of Rock (2003), Kung Fu Panda (2008), Miss Congeniality (2000), Ninotchka (1939)\\
\midrule
Funny& Ace Ventura: When Nature Calls (1995), Ace Ventura: Pet Detective (1994), Hot Shots! Part Deux (1993), Army of Darkness (1993), South Park: Bigger, Longer and Uncut (1999), Auntie Mame (1958), Blazing Saddles (1974), Clerks (1994), Grand Day Out with Wallace and Gromit, A (1989), Hitchhiker's Guide to the Galaxy, The (2005) & Spanish Prisoner, The (1997), Son of Dracula (1943), Ghost Dog: The Way of the Samurai (1999), Ferngully: The Last Rainforest (1992), High Crimes (2002), Cadillac Man (1990), Bad Boys II (2003), House of Wax (1953), Fire in the Sky (1993), Step Up 2 the Streets (2008)\\
\midrule
Population& India, Nigeria, People's Republic of China, Iran, Pakistan, United States of America, Russia, Indonesia, Egypt, Bangladesh & Dominica, Nauru, Andorra, Cook Islands, Saint Vincent and the Grenadines, Seychelles, Palau, Northern Mariana Islands, Liechtenstein, Niue\\
\bottomrule
\end{tabular}
\caption{We show the top and bottom ranked entities for five features: \emph{sweetness} and \emph{saltiness} from the \textit{food} dataset, \emph{scary} and \emph{funny} from movies, and \emph{countries population} from WD2. Results were obtained with the pairwise Llama2-13B model trained on WD+TG+Rocks for \emph{sweetness} and \emph{saltiness}, trained on WD+Taste+Rocks for \emph{scary} and \emph{funny}, and trained on WD1 for \emph{population}. \label{tabLLaMA-2Qualitative}}
\end{table*}

\end{document}